\documentclass[conference]{IEEEtran}
\usepackage{times}

\usepackage[numbers]{natbib}
\usepackage{multicol}
\usepackage[bookmarks=true]{hyperref}
\usepackage{graphicx} 
\usepackage{wrapfig}

\usepackage[export]{adjustbox}
\usepackage[position=top]{subfig}
\usepackage{subcaption} 

\usepackage{threeparttable}
\usepackage{makecell}
\usepackage{multirow}
\usepackage{booktabs} 
\usepackage{array}
\usepackage{pifont}
\usepackage{float}
\usepackage{tabularx}
\usepackage[ruled, lined, longend, linesnumbered]{algorithm2e}
\usepackage{amsmath}
\usepackage{amssymb}
\usepackage{amsfonts}
\usepackage{amsthm}
\let\labelindent\relax
\usepackage{enumitem}
\usepackage{thmtools}
\usepackage{thm-restate}

\usepackage{times}
\usepackage[numbers]{natbib}
\usepackage{multicol}
\usepackage{tikz}    %
\usepackage{graphicx}
\usepackage[bookmarks=true]{hyperref}
\usepackage{multirow}
\usepackage{tcolorbox}
\usepackage{balance}
\usepackage{tcolorbox}

\newtheoremstyle{mydef}%
{3pt}{3pt}
{\normalfont}
{}
{\bfseries}
{.}
{0.5em}
{\thmname{#1}\thmnumber{ #2}\thmnote{ (#3)}}

\theoremstyle{mydef}

\usepackage{tcolorbox}
\tcbuselibrary{theorems}

\newtheorem{definition}{Definition}

\newtheorem*{theorem*}{Theorem}

\usepackage{booktabs}
\usepackage{multirow}

\usepackage{tcolorbox}
\tcbuselibrary{breakable,skins}

\newtcolorbox{examplebox}[2][]{
  enhanced,
  breakable,
  colframe=black,
  colback=white,
  boxrule=0.6pt,
  arc=2mm,
  left=1.2mm,right=1.2mm,top=1.2mm,bottom=1.2mm,
  fontupper=\small\sffamily,
  fonttitle=\bfseries\sffamily\color{white},
  colbacktitle=black,
  title=#2,
  width=\columnwidth,
  #1
}

\newcommand{\mypara}[1]{\smallskip\noindent\textbf{#1}~}

\begin{document}

\title{SkillVLA: Tackling Combinatorial Diversity in Dual-Arm Manipulation via Skill Reuse}

\author{Xuanran Zhai$^{*1}$, Zekai Huang$^{*1}$, Longyan Wu$^{*3,6}$, Qianyou Zhao$^5$, Qiaojun Yu$^4$, Jieji Ren$^5$,\\ Ce Hao$^{\dagger1,2}$, and Harold Soh$^{1,7}$\\
$^{1}$National University of Singapore, $^{2}$Beijing Zhongguancun Academy, $^{3}$Shanghai Innovation Institute,\\ $^{4}$Shanghai AI Laboratory, $^{5}$Shanghai Jiao Tong University, $^6$Fudan University, $^7$Smart Systems Institute, NUS\\
$^*$ Equal contribution. $^\dagger$ Corresponding author: \href{cehao@u.nus.edu}{\texttt{cehao@u.nus.edu}}}


%

\twocolumn[{%
	\renewcommand\twocolumn[1][]{#1}%
	\maketitle
        \vspace{-5mm}
	\begin{center}
		\includegraphics[width=0.95\textwidth]{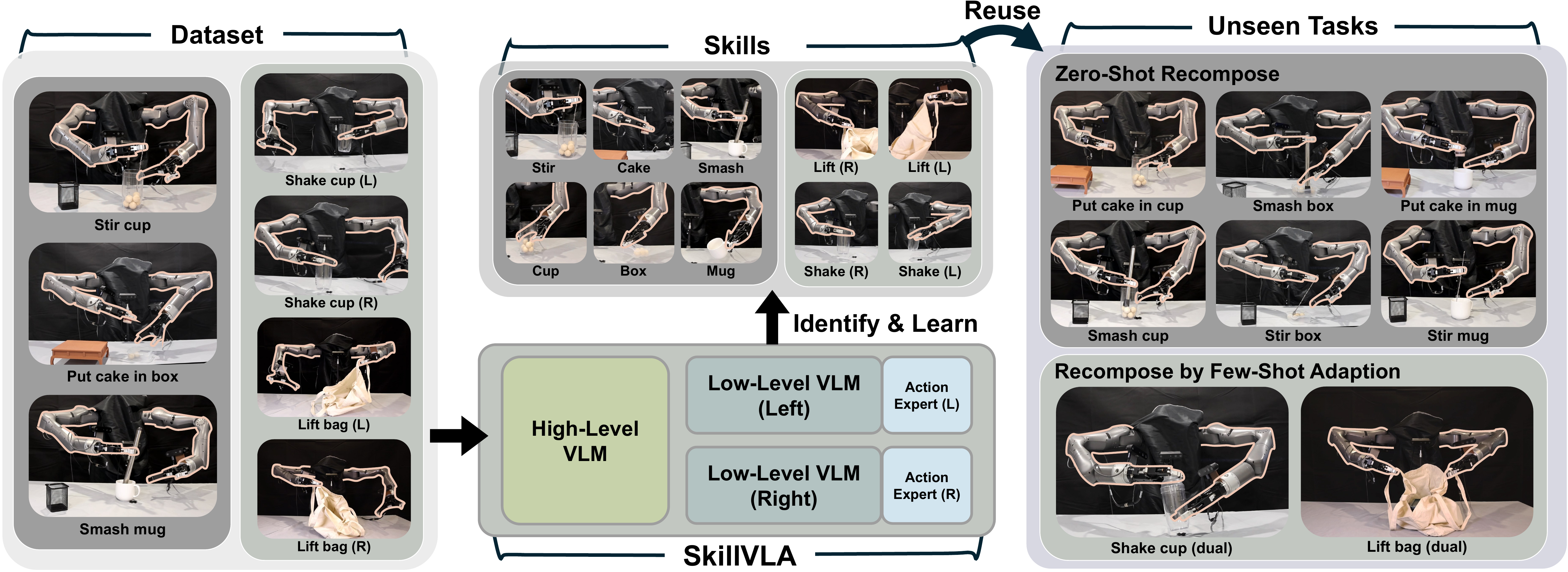}
		\captionof{figure}{\small \label{fig: tsr} SkillVLA extracts single-arm skills from training data with hierarchical reasoning and skill-adaptive learning, being able to recompose them into unseen combinations during test time.
        } 
	\end{center}
}]
\begin{abstract}
Recent progress in vision–language–action (VLA) models has demonstrated strong potential for dual-arm manipulation, enabling complex behaviors and generalization to unseen environments. However, mainstream bimanual VLA formulations largely overlook the critical challenge of \emph{combinatorial diversity}. Different pairings of single-arm behaviors can induce qualitatively distinct task behaviors, yet existing models do not explicitly account for this structure. We argue that effective bimanual VLAs should support skill reuse---the ability to recombine previously learned single-arm skills across novel left–right pairings---thereby avoiding the need to separately learn every possible combination. Current VLA designs entangle skills across arms, preventing such recomposition and limiting scalability.
To address this limitation, we propose SkillVLA, a framework explicitly designed to enable skill reuse in dual-arm manipulation. Extensive experiments demonstrate that SkillVLA substantially improves skill composition, increasing overall success rate from $0\%$ to $51\%$, and achieves strong performance on cooperative and long-horizon tasks.
\end{abstract}
\IEEEpeerreviewmaketitle

\section{Introduction}~\label{Sec: Intro}
Bimanual manipulation expands a robot's effective workspace and enables concurrent, coordinated actions, supporting a wider range of tasks than single-arm systems~\cite{dualarmsurvey,yoto,xie2020deep, chitnis2020efficient,chen2025robotwin,KimJ2-RSS-25}. 
In this context, recent vision--language--action (VLA) models~\cite{pi0,pi05,rdt,rdt2,LLaVLA,dexvla} have shown impressive performance on bimanual manipulation; by jointly predicting both arms' actions (e.g., via a concatenated action vector) with large-capacity architectures, VLA methods can represent complex coordinated behaviors. Further, by leveraging vision--language pretraining of VLM models, recent bimanual VLAs have shown the ability to generalize to novel scenes and instructions~\cite{pi05,internvla,pi06,qwen25vl7b,PaliGemma,qwen3vl,pi0,kim2024openvla}.


Despite recent progress, learning generalist bimanual policies remains challenging, as dual-arm control introduces richer behaviors patterns. Specifically, real-world bimanual behaviors exhibit pronounced \textit{\textbf{combinatorial diversity}}. As shown in Fig.~\ref{fig: tsr}, many dual-arm tasks can be viewed as compositions of single-arm behaviors, and different pairings of left- and right-arm skills induce different bimanual tasks. As the underlying skill set grows, the number of possible pairings increases quadratically, resulting in an overwhelming number of combinations corresponding to different tasks.
Current VLA formulations largely overlook combinatorial diversity. Predicting both arms via action concatenation requires the model to learn the joint distribution of left- and right-arm distributions. While this design is simple and captures tight coordination, it confines outputs to action pairings observed in demonstrations. As a result, the policy has limited combinatorial generalization and cannot form new bimanual behaviors by recombining single-arm skills, which makes them unable to address combinatorial diversity. We refer to this limitation as \emph{\textbf{skill entanglement}} and analyze it in detail in Sec.~\ref{subsec:vla_skill_reuse_problem}.

To address the combinatorial diversity, a model must be able to effectively reuse robot skills learned from given demonstrations in a more flexible and thorough way. We name this ability as \textbf{\textit{skill reuse}}. Concretely, instead of mechanically treating every bimanual behavior as indivisible dual-arm primitive, an ideal model should automatically uncover the latent skill structure within demonstrations and perform reasoning at the skill level. For bimanual behaviors that are essentially composition of single-arm skills, the model should reuse the discovered skills across different left–right combinations and compose them to form novel pairings that are not explicitly observed during training. We view such flexible reuse as a key foundation for broad generalization in bimanual settings, and provide a more formal definition and discussion in Sec.~\ref{Sec:skill_reuse}.

We propose \textbf{SkillVLA} to enable skill reuse and address combinatorial diversity. Unlike conventional VLAs relying on monolithic dual-arm reasoning throughout, SkillVLA automatically identifies skill structures from demonstrations and performs \textit{skill-adaptive} action generation. Concretely, for single-arm skills, it can retrieve and invoke them as independent units for recomposition or further adaptation. For behaviors that require collaboration, the model instead captures inter-arm dependence.
SkillVLA achieves this with a two-level reasoning architecture. A shared high-level VLM, equipped with a collaboration estimator, captures global task intent and makes skill decisions. At the lower level, the policy follows the high-level instruction and switches between operation modes: it performs per-arm reasoning and action generation for single-arm skills, while allowing these two streams to be entangled by communication when cooperative behavior is deemed necessary by the high-level reasoning. Through skill-adaptive generation and per-arm reasoning, SkillVLA further exploits the VLM’s capability by extending reasoning to a finer semantic granularity, leading to stronger generalization.

We conduct real-robot experiments on 20 manipulation tasks and two long-horizon multi-stage tasks to validate SkillVLA's skill reuse and generalization ability.
First, to evaluate if models are able to tackle combinatorial diversity, we test them on unseen tasks composed by learned skills, where strong bimanual VLA baselines achieve near-zero success due to skill entanglement. SkillVLA raises the success rate to \(\mathbf{51\%}\), demonstrating a decisive improvement in combinatorial generalization.
Second, to confirm that our method does not sacrifice inter-arm cooperation, we evaluate on three highly-cooperative tasks, where SkillVLA matches the performance of the baselines, showing that its adaptive communication remains expressive for close coordination.
Third, we study two multi-stage long-horizon tasks that interleave collaborative and independent phases. While SkillVLA correctly identifies cooperation level required and, further, completes the tasks faster by recomposing skills to parallelize the two arms when possible, reducing execution time by \(\mathbf{21\%}\).
Finally, we conduct a continual-learning study to show that skill reuse also benefits long-term skill learning: SkillVLA can effectively reuse existing  behaviors for acquiring new skills, achieving significantly better performances under limited demonstrations.

In summary, this paper makes three main contributions:
\begin{itemize}[noitemsep]
    \item We identify the \emph{skill entanglement} issue in current VLAs and formulate the bimanual
    \emph{skill reuse} problem.
    \item We propose SkillVLA, which enables flexible skill reuse to tackle combinatorial diversity.
    \item We validate SkillVLA on a real dual-arm robot, demonstrating combinatorial generalization ability via skill reuse, while maintaining strong performance on cooperative and long-horizon tasks.
\end{itemize}

\section{Related Work}~\label{Sec: Related Works}
Dual-arm manipulation has long been a central topic in robotics. Early work largely focused on methods specialized for particular task families, such as assembly~\cite{forcevla,Zhao-RSS-23,ge2025dual,ALOHA}, cloth folding~\cite{maitin2010cloth,bersch2011bimanual,weng2022fabricflownet,salhotra2023learning,canberk2022cloth,avigal2022speedfolding,chi2024universal}, and bagging~\cite{autobag,bahety2023bag,chen2023bagging}. To improve adaptability, subsequent efforts introduced platforms with richer actuation, including dexterous end-effectors~\cite{dexcap,shaw2024bimanual,Bunny-VisionPro} and mobile bases~\cite{equibot,fu2024mobile}. More recently, motivated by the goal of generalist bimanual policies, several works have scaled training data by learning from human demonstration videos~\cite{yoto,grauman2022ego4d,papagiannis2025r}.

Generalist bimanual policies, however, remained limited until the emergence of VLA models~\cite{driess2023palm,tinyvla,rdt,LLaVLA,rt2,3dvla,pi0} accelerated progress in this direction. These approaches typically train end-to-end imitation learning policies on top of pretrained vision--language models. Benefiting from large model capacity and large-scale training on diverse robot datasets~\cite{oxedataset,DROID,shafiullah2023bringing,lin2025data,octopi}, they have shown improved environmental generalization, including better robustness to complex or unseen scenes. Nonetheless, generalization across diverse tasks remains challenging. Building on this line of work, we propose SkillVLA to improve action-level generalization.

\section{Problem Formulation}
\label{Sec: Problem}
We consider a bimanual robot with left and right manipulators.
At each time step, the robot receives an input \(x \in \mathcal{X}\) (usually observation, instruction and robot state) and produces actions
\(a_L \in \mathcal{A}_L\) and \(a_R \in \mathcal{A}_R\); we write \(a=(a_L,a_R)\).
We are given a dataset of demonstrations
\(\mathcal{D} = \{(x_i, a_L^i, a_R^i)\}_{i=1}^\mathcal{I}\),
where each tuple is a state--action sample extracted from expert trajectories.
We assume these expert actions are generated by executing skills drawn from an underlying skill set
\[
S = S_{\mathrm{pair}} \cup S_D, \ 
S_{\mathrm{pair}} = \{(s_L^{m}, s_R^{m})\}_{m=1}^{M}, \ S_D=\{s_D^n\}_{n=1}^N
\]
where \(S_L\) and \(S_R\) are the single-arm skill sets for the left and right arm
(\textit{Def.}~\ref{def:single-arm-skill}), and
\(S_D\) is a set of genuinely dual-arm (bimanual) skills (\textit{Def.}~\ref{def:bimanual-skill}).
Here \(S_{\mathrm{pair}}\) collects the \emph{paired} single-arm skills that actually appear in the demonstrations.

Our goal is to learn a bimanual policy \(\pi_\theta(a_L, a_R \mid x)\) that is able to effectively
\textit{\textbf{reuse}} these learned skills (\textit{Def.}~\ref{def:skill-mastery}),
which we view as a key factor for resolving combinatorial diversity building generalist bimanual manipulation policies.
\section{Skill Reuse in Dual-Arm Manipulation}
\label{Sec:skill_reuse}

In this section, we first give specific definitions of skills and skill reuse (Sec.~\ref{subsec:skill_defs}),
then we analyze what is required to solve this problem: for skill reuse to be possible, the model must
(i) select suitable skills for any given scene \(x\) (Sec.~\ref{subsec:skill_selector}), and
(ii) generate correct actions for the selected skill (or skill pair).
The latter motivates explicitly distinguishing between single-arm and dual-arm skills during training and execution,
which current VLA-based methods do not support (Sec.~\ref{subsec:vla_skill_reuse_problem}).
\subsection{Definitions}
\label{subsec:skill_defs}
To formulate and analyze skill reuse in a general bimanual setting, we first define what we mean by a \emph{skill}.
We distinguish between \emph{single-arm skills}, which can be executed by one arm in isolation, and
\emph{dual-arm skills}, which require coordinated behavior between the two arms.
\begin{definition}[Single-Arm Skills]
\label{def:single-arm-skill}
For \(\kappa \in \{L,R\}\), $S_\kappa$ is the set of all possible left/right arm skills. A single-arm skill \(s_\kappa \in S_\kappa\) is a conditional distribution \(\pi_{s_\kappa}(a_\kappa \mid x)\).
For any pair \((s_L,s_R) \in S_L \times S_R\), their  composition is defined as
\[
\pi_{s_L,s_R}(a_L,a_R \mid x)
= \pi_{s_L}(a_L \mid x)\,\pi_{s_R}(a_R \mid x).
\]
\end{definition}
\begin{definition}[Dual-Arm Skills]
\label{def:bimanual-skill}
A dual-arm skill \(s_D \in S_D\) is a joint conditional distribution
\(\pi_{s_D}(a_L, a_R \mid x)\) such that the conditional mutual information
\(I_{\pi_{s_D}}(a_L; a_R \mid x) > 0\).
Equivalently, \(\pi_{s_D}(a_L, a_R \mid x)\neq\pi_{s_L}(a_L \mid x)\,\pi_{s_R}(a_R \mid x)\) for any
single-arm skills \(s^L,s^R\).
\end{definition}
Achieving inter-arm coordination for dual-arm skills requires an information pathway that induces dependence between \(a_L\) and \(a_R\). We denote this pathway conceptually by an inter-arm message \(m\), yielding action generation forms such as \(\pi_L(a_L \mid x, Y_L, m_L)\) and \(\pi_R(a_R \mid x, Y_R, m_R)\). In practice, inter-arm message may be realized in various ways, such as explicit message passing, or implicitly through shared parameters as in common monolithic policies.

For any possible scene \(x\), we assume there exists a (possibly multi-valued) set of \emph{correct skills}
\(S^\star(x) \subseteq (S_L \times S_R) \cup S_D\) that are appropriate for that scene.
We decompose this set as $S^\star(x)=S^\star_{\mathrm{pair}}(x)\cup S^\star_D(x)$, where
\[
S^\star_{\mathrm{pair}}(x) = S^\star(x) \cap (S_L \times S_R),
\qquad
S^\star_D(x) = S^\star(x) \cap S_D,
\]
corresponding to correct skills that can be expressed as a composition of single-arm skills and correct skills that are
genuinely dual-arm.
Note that \(S^\star_{\mathrm{pair}}(x)\) is defined over the full product \(S_L \times S_R\),
and is not restricted to the paired set \(S_{\mathrm{pair}}\); in particular, it may contain recombinations of
single-arm skills that never co-occur in the demonstrations.
\begin{definition}[Skill Reuse]
\label{def:skill-mastery}
For a bimanual policy learned from skill set $S=S_{\text{pair}}\cup S_D=\{(s_L^{m},s_R^{m})\}\cup\{s_D^n\}$, we define its possible behavior set as $S^p=(\{s_L^{m}\}\times \{s_R^{m}\})\cup \{s_D^n\}$.
We say that the policy exhibits \emph{\textbf{skill reuse}} if, for every scene \(x\) (possibly unseen) with
\(S^\star(x) \cap S^p \neq \emptyset\), there exists a skill
\(s \in S^\star(x) \cap S^p\) such that \(\pi_\theta(a_L,a_R \mid x) \approx \pi_s(a_L,a_R \mid x)\).
\end{definition}
Equivalently, this requirement decomposes into two major subproblems:
(i) \emph{single-arm composition}: if the correct behavior for \(x\) is a composition of single-arm skills, i.e.
\(S^\star(x)\cap S^p = \{(S_L^i,S_R^j)\}\) for some $i, j$, the policy must be able to realize this composition, even for \((s^i_L,s^j_R)\) that never appeared in
\(S_{\mathrm{pair}}\);
and (ii) \emph{dual-arm reproduction}: when the correct skill is genuinely dual-arm, i.e.
\(S^\star(x)\cap S^p = \{s_D^i\}\) for some $i$, the policy should also produce action for $s_D^i$ while maintaining its inter-arm coordination structure. For cases where multiple skill choices are valid, specifically, the model should then select and produce one of them.

\mypara{Remark (Fast bimanual continual learning by skill reuse).}
In practice, many dual-arm skills are close to simple compositions of two single-arm skills, in the sense that each arm
largely follows a independent per-arm behavior under the same context \(x\).
The challenge is the coordination between the two arms: the joint
action distribution deviates from an independent product due to inter-arm coupling (cf.\ \(I(a_L;a_R\mid x)>0\) in
Def.~\ref{def:bimanual-skill}).
Therefore, if a model can flexibly invoke reusable single-arm skills, acquiring a new bimanual skill can often be
done with minimal finetuning by mainly learning additional information for coupling on top of the existing per-arm
skills, rather than relearning both arms from scratch.
Therefore, effective skill reuse can  improve efficiency in continual learning or large-scale learning.

\subsection{Skill Selector}
\label{subsec:skill_selector}
Given a library of skills, action generation can be viewed as first selecting which skill (or skill
pair) to use for a given scene \(x\), and then sampling actions from the corresponding skill distributions.
We assume a (finite) skill library consisting of indexed single-arm skills
\(\{s_L^k\}_{k \in K_L} \subseteq S_L\) and \(\{s_R^l\}_{l \in K_R} \subseteq S_R\),
and indexed dual-arm skills \(\{s_D^d\}_{d \in K_D} \subseteq S_D\).
We introduce a discrete \emph{skill index} random variable
\[
Y \in \mathcal{Y} := K_D \,\cup\, (K_L \times K_R),
\]
where \(Y = d \in K_D\) means that the dual-arm skill \(\pi_{s_D}(a_L,a_R \mid x)\) should be used, and
\(Y = (k,l) \in K_L \times K_R\) means that the single-arm skills
\(\pi_{s_L^k}(a_L \mid x)\) and \(\pi_{s_R^l}(a_R \mid x)\) should be used concurrently.

An ideal high-level selector defines a distribution \(p^\star(Y \mid x)\), and the resulting action distribution is the
mixture
\begin{align*}
\pi^\star&(a_L,a_R \mid x)
=
\sum_{d \in K_D} p^\star(Y = d \mid x)\,\pi_{s_D}(a_L,a_R \mid x)
\;\\&+
\sum_{(k,l) \in K_L \times K_R} p^\star(Y = (k,l) \mid x)\,
\pi_{s_L^k}(a_L \mid x)\,\pi_{s_R^l}(a_R \mid x).
\end{align*}
We use the term \emph{skill selector} for this conceptual mechanism, without assuming a specific module or
architecture.
Ideally, the skill selector should select appropriate skills not only for scenes seen in the demonstrations, but
also for novel inputs where the correct skill configuration has never been observed.

\subsection{Skill Entanglement in Current VLAs}
\label{subsec:vla_skill_reuse_problem}

VLAs are often built upon a pretrained VLM, which provides strong generalization over visual scenes and natural-language instructions. 
Typically, a VLA has a additional action module (or ``action expert'') to generate actions. In bimanual manipulation, the action is commonly represented as a single vector by concatenating the left- and right-arm actions.

The VLM is a natural candidate
to implement a \emph{generalizable skill selector}---i.e., to map a scene \(x\) to an appropriate skill index \(Y\) (or an
equivalent decision variable) that can generalize beyond the demonstrated scenes.
However, even if the upstream skill decision is sufficiently able to
separate scenes requiring different skills, can the downstream action generation mechanism \emph{reuse} skills in the sense of Sec.~\ref{subsec:skill_defs}?

We argue that common VLA designs exhibit two forms of \textit{skill entanglement} that hinder effective skill reuse:

\mypara{Action Entanglement.}
Many bimanual VLA policies are trained to predict a single concatenated joint action vector \((a_L,a_R)\). This monolithic supervision couples the two arms at the output level and encourages the model to fit the empirical \emph{joint} distribution induced by paired demonstrations. Consequently, the learned policy may internalize dataset-specific cross-arm correlations rather than isolating reusable single-arm structure.
This becomes problematic for skill reuse and recomposition. Even if the upstream vision--language reasoning can distinguish scenes requiring different skills, the downstream action generator may fail to (i) disentangle single-arm skills from dual-arm coordination patterns and (ii) support recomposition of single-arm skills beyond the left--right pairings observed during training. In other words, joint-action learning can bias the model toward reproducing demonstrated bimanual patterns, limiting its ability to generalize to unseen combinations of per-arm behaviors.

\mypara{Latent entanglement in action-expert VLAs.}
As previously mentioned, recent VLA methods augment a pretrained VLM with a dedicated action-generation module 
(e.g., $\pi_0$/$\pi_{0.5}$~\cite{pi0,pi05}, RDT2~\cite{rdt2}, DexVLA~\cite{dexvla}).
Abstractly, a VLM encodes context $x$ into a representation $z$, and an action module predicts bimanual actions conditioned on $z$,
\begin{equation*}
p(a_L,a_R \mid x)
= \int p(a_L,a_R \mid z)\, p(z \mid x)\, dz,
\end{equation*}
where \(p(z \mid x)\) is induced by the VLM backbone and \(p(a_L,a_R \mid z)\) is implemented by the action expert.

While this architecture can be effective in practice, it introduces an additional pathway for skill entanglement. In bimanual imitation, the shared latent \(z\) learned from paired demonstrations may implicitly encode cross-arm dependencies. This \emph{latent entanglement} can degrade skill recomposition when the policy is evaluated on unseen left--right pairings, since the action expert conditions both arms on a representation that already mixes information across arms.

\begin{figure*}[t]
    \centering
    \includegraphics[width=0.90\linewidth]{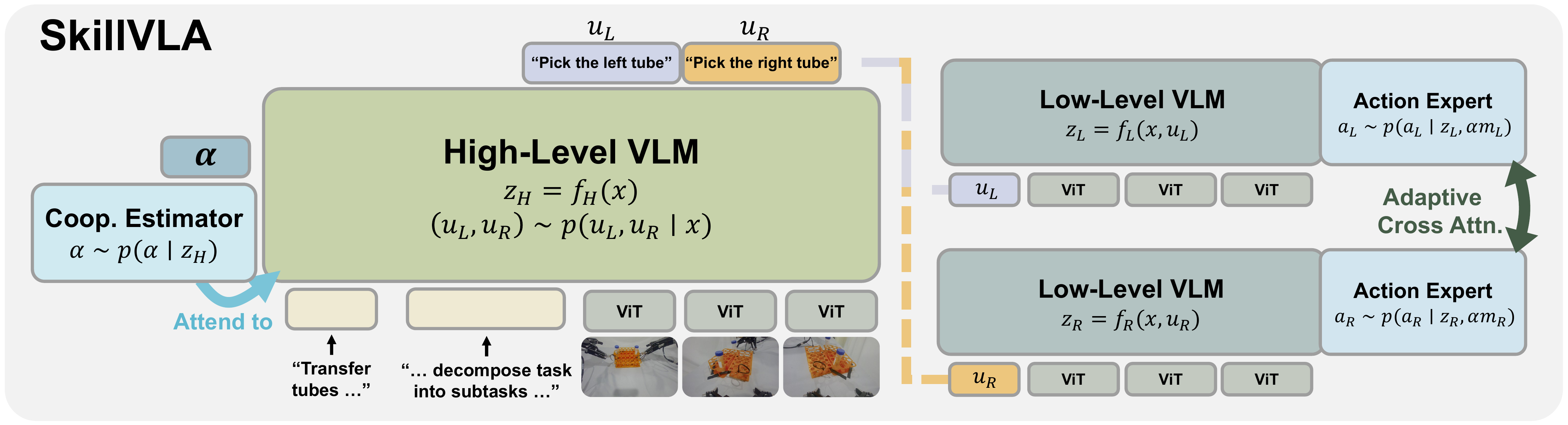}
    \caption{\small \textbf{SkillVLA framework.} SkillVLA adopts a two-level reasoning pipeline, where the high-level VLM generates separate subtasks for arms and low-level VLMs further process the prompts to instruct action generation. Inter-arm cross-attention enables cooperative behaviors generation, controlled by a collaboration estimator that identifies the operation mode required.}
    \label{Fig:pipeline}
    \vspace{-3.8mm}
\end{figure*}

\section{Method: SkillVLA}~\label{Sec: Method}
In this section, we present \textbf{SkillVLA}, a method designed to enable effective skill reuse to address combinatorial diversity and accelerate the acquisition of new skills.
We first describe the overall architecture, which realizes skill reuse via a two-level reasoning pipeline and a collaboration estimator for behavior-type identification.
We then introduce additional training strategies that help the model better capture skill structure and infer the degree of inter-arm cooperation.

\subsection{Method Pipeline}
~\label{Subsec: method1}
Our method (overview in Fig.~\ref{Fig:pipeline}) follows the common VLA paradigm, with a top-level VLM and actions generated via an iterative flow-matching process~\cite{lipman2023flow,liu2023flow}. 
In our implementation, we use the pretrained PaliGemma~\cite{PaliGemma} backbone released with $\pi_{0.5}$~\cite{pi05} for initiating the VLMs.
Our method consists of two main functional components, described below:

\mypara{Two-level reasoning (skill selection and action generation).}
Because an explicit skill library is typically unavailable in practice, we aim for the model to \emph{discover} and \emph{instantiate} skill representations that support both learning and reuse. Skills can be represented in many forms; in {SkillVLA} we use natural language as the skill descriptor, which aligns naturally with the VLM backbone. We implement this choice via a two-level reasoning pipeline.

As shown in Fig.~\ref{Fig:pipeline}, the high-level module explicitly produces per-arm sub-prompts that act as skill descriptors. Concretely, we prompt the VLM to generate two texts \((u_L,u_R)\) in a fixed format:
\[
(u_L,u_R) \sim p(u_L,u_R \mid x),
\]
where \(u_L\) and \(u_R\) describe what the left and right arms should do, respectively (e.g., ``pick the cake on the right'' vs.\ ``pick the box on the left''). This representation targets task intent and \emph{explicitly} decouples per-arm skill selection, enabling flexible single-arm recomposition: novel combinations can be formed by pairing previously generated (or learned) \(u_L\) and \(u_R\) in new scenes. During low-level skill learning, we freeze the high-level VLM to preserve its vision--language generalization while training the action components.

At the low level, per-arm actions are generated by two separate streams. Each stream uses its own low-level VLM (fine-tuned separately, e.g., via LoRA~\cite{lora}) to process the visual input and the corresponding per-arm prompt, producing a per-arm latent representation \(z_i = f_i(x,u_i)\), where \(i \in \{L,R\}\). Action experts then predict actions conditioned on the corresponding latent and the arm state. To support coordinated bimanual behaviors when needed, we introduce an adaptive cross-attention mechanism between the action experts to capture inter-arm dependence:
\[
a_i \sim p\!\left(a_i \mid z_i,\, \alpha\, m_i\right),
\]
where \(m_i\) is the inter-arm message produced by cross-attention. The \emph{cooperation level} signal \(\alpha\) gates this message to enable skill-adaptive action generation and is further elaborated upon below.

\mypara{Cooperation estimator (behavior mode identification).}
While inter-arm communication is useful for capturing low-level dependence, it should be \emph{selectively} enabled; for single-arm skills, the two arms should remain largely disentangled during both training and evaluation. To this end, we introduce a cooperation estimator that attends to the high-level VLM representation and predicts a scalar \(\alpha \in [0,1]\), indicating the degree of inter-arm cooperation (larger \(\alpha\) implies stronger coupling). Formally, the estimator predicts
\[
\alpha \sim p(\alpha \mid z_H), \qquad z_H = f_H(x),
\]
where \(z_H\) is the latent representation produced by the frozen high-level VLM. This signal serves as a mode identifier, indicating whether the current behavior is better explained as (i) a composition of single-arm skills or (ii) a cooperative dual-arm skill. We gate inter-arm message passing by \(\alpha\), enabling the policy to interpolate between per-arm generation and coupled dual-arm generation.

To train \(\alpha_t\), we use a simple \emph{communication usefulness} objective derived from behavioral cloning (BC). At timestep \(t\), the BC loss is
\[
\mathcal{L}_{\mathrm{BC}}
=
\|\hat a_t^L-a_t^L\|_2^2
+
\|\hat a_t^R-a_t^R\|_2^2.
\]
We compute \(\mathcal{L}_{\mathrm{BC}}^{\mathrm{on}}\) with cross-attention enabled and \(\mathcal{L}_{\mathrm{BC}}^{\mathrm{off}}\) with cross-attention disabled. We then define
\[
\mathcal{L}_{\mathrm{coop}}
=
\lambda \cdot
\Big(\mathcal{L}_{\mathrm{BC}}^{\mathrm{on}} - \mathcal{L}_{\mathrm{BC}}^{\mathrm{off}}\Big)_{\mathrm{sg}}
\cdot \alpha_t,
\]
where \((\cdot)_{\mathrm{sg}}\) denotes stop-gradient. Minimizing \(\mathcal{L}_{\mathrm{coop}}\) increases \(\alpha_t\) when communication reduces the BC error and decreases \(\alpha_t\) otherwise.

\subsection{Additional Cooperation-Level Learning}
\label{Subsec:cooperation_learning}
Because \(\alpha\) directly gates cross-arm interaction, accurately inferring the cooperation level is critical. We  introduce additional mechanisms to encourage reliable cooperation estimation, which we use by default in our implementation.

\mypara{Priors and regularizers for cooperation learning.}
VLMs are pretrained on datastes and thus encode broad task semantics and commonsense regularities (e.g., when two arms are typically required). This makes them a natural source of priors for estimating task-dependent cooperation.
To distill this information into a lightweight estimator, we use an off-the-shelf VLM to produce a prior coordination strength \(\alpha^{\mathrm{vlm}} \in [0,1]\) for the current scene and task (or \(\alpha^{\mathrm{vlm}}\in\{0,1\}\) when using a discrete gate).

Let \(\tilde\alpha_t\) denote the gate value actually used for communication (i.e., the continuous \(\alpha_t\), or the soft discrete probability \(\hat y_t\) defined below). We regularize \(\tilde\alpha_t\) toward the VLM prior and further encourage (a) {sticky} gating for temporal smoothness and (b) conservative communication, keeping the gate small unless interaction is beneficial:
\begin{gather*}
\mathcal{L}_{\mathrm{prior}}
=
\ell(\tilde\alpha_t,\alpha^{\mathrm{vlm}}),\\
\mathcal{L}_{\mathrm{sticky}}
=
\ell(\tilde\alpha_t,\tilde\alpha_{t-1}),
\qquad
\mathcal{L}_{\mathrm{sup}}
=
\tilde\alpha_t.
\end{gather*}
We set \(\ell(u,v)=\|u-v\|_2^2\) when \(\tilde\alpha_t\) is continuous, and use a Bernoulli cross-entropy
\(\ell(u,v)=-v\log u-(1-v)\log(1-u)\) when \(\tilde\alpha_t\in(0,1)\) is interpreted as a probability.

\begin{figure*}[t]
    \centering
    \includegraphics[width=0.90\linewidth]{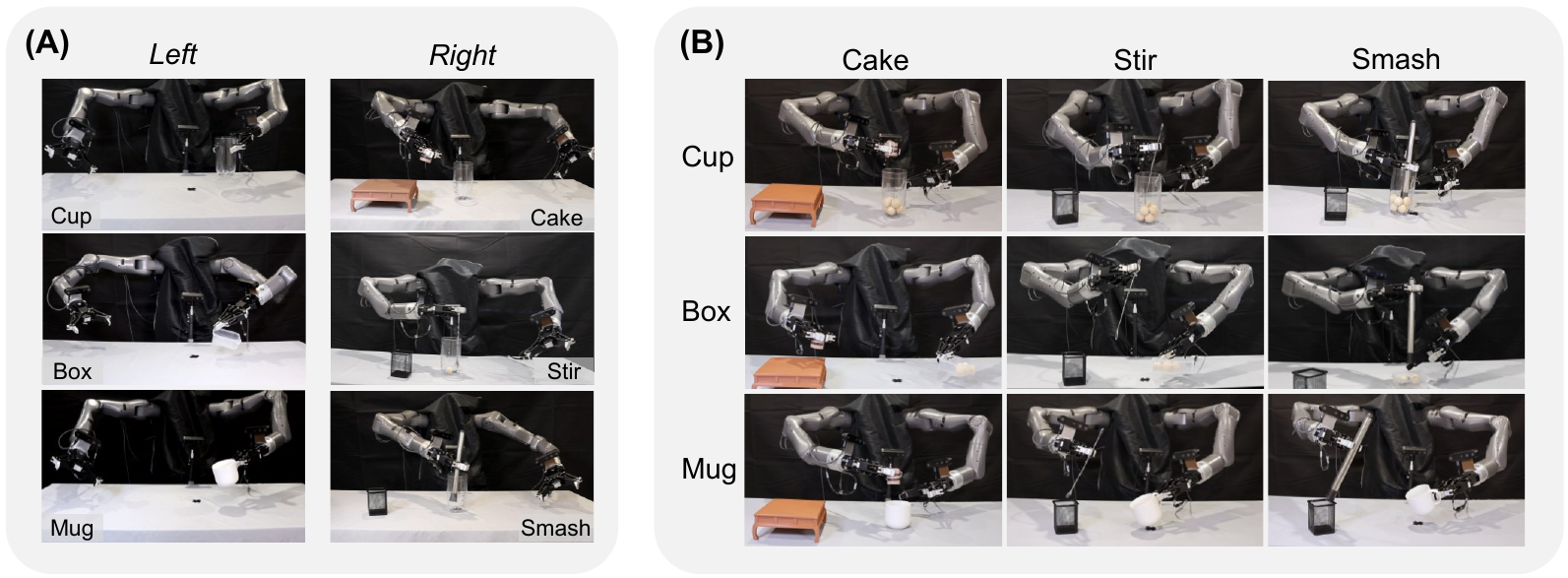}
    \caption{\small \textbf{Skill recomposition tasks.} \textbf{(A):} The models are trained on demonstrations of three skills for each arm. \textbf{(B):} After the models have learned the skills, zero-shot tests are conducted for every possible combinations of left and right-arm skills.}
    \label{Fig:exp_real_recomp}
    \vspace{-2mm}
\end{figure*}
\begin{table*}[t]
    \centering
    \caption{\small Success Rates on Skill Recomposition Tasks $\uparrow$ }
    \label{Tab: exp_results_recompo}
    \renewcommand\arraystretch{1.2}
    \resizebox{0.98\textwidth}{!}{
    \begin{threeparttable}
    \begin{tabular}{l|cccccccccc}
        \toprule
        \multirow{2}{*}{Methods} & \multicolumn{10}{c}{Recomposition Tasks} \\
        & Cup$\times$Cake & Cup$\times$Stir & Cup$\times$Smash 
        & Box$\times$Cake & Box$\times$Stir & Box$\times$Smash
        & Mug$\times$Cake & Mug$\times$Stir & Mug$\times$Smash & Avg.\\  
        \midrule
        $\pi_{0.5}$   
        & 0.0 & 0.0 & 0.0
        & 0.0 & 0.0 & 0.0 
        & 0.0 & 0.0 & 0.0 & 0.0 \\
        $\pi_{0}$-FAST 
        & 0.0 & 0.0 & 0.0
        & 0.0 & 0.0 & 0.0 
        & 0.0 & 0.0 & 0.0 & 0.0 \\
        TwinVLA      
        & 0.1 & 0.0 & 0.0
        & 0.2 & 0.0 & 0.1 
        & 0.0 & 0.0 & 0.0 & 0.04 \\
        SkillVLA
        & \textbf{0.7} & \textbf{0.4} & \textbf{0.5} 
        & \textbf{0.6}& \textbf{0.4}&  \textbf{0.5} 
        & \textbf{0.6} & \textbf{0.3} & \textbf{0.6} &\textbf{0.51}\\
        \bottomrule
    \end{tabular}
\end{threeparttable}
}
\vspace{-3mm}
\end{table*}
\begin{table}[t]
\centering
\caption{\small Success Rates on Skills Learned}
\begin{tabular}{l|ccccccc}
\toprule
\multirow{2}{*}{Methods} & \multicolumn{7}{c}{Skills Learned} \\
& Cup & Box & Mug & Cake & Stir & Smash & Avg.\\
\midrule
$\pi_{0.5}$ & 0.7& \textbf{1.0}& \textbf{0.8}& 0.8& \textbf{0.6}& 0.7& 0.77\\
$\pi_{0}$-FAST & \textbf{0.9}& 0.8& 0.7& 0.9& 0.5& 0.4& 0.70\\
TwinVLA    & 0.7&0.8& 0.5& 0.8& 0.5& 0.7& 0.67\\
SkillVLA   & 0.8& 0.7& \textbf{0.8}& \textbf{1.0} & \textbf{0.6}& \textbf{0.8}& \textbf{0.78}\\
\bottomrule
\end{tabular}
\label{Tab: exp_results_skills_learned}
\vspace{-3mm}
\end{table}

\mypara{Cooperation level discretization.}
In practice, a continuous gate \(\alpha_t\) can exhibit small but persistent fluctuations that destabilize action generation. To improve stability, we (optionally) discretize the gate by restricting \(\alpha_t \in \{0,1\}\).
Specifically, the model predicts \(\hat y_t \in (0,1)\), the probability of enabling cross-arm communication, and we train it with a binary cross-entropy loss:
\[
\mathcal{L}_{\mathrm{disc}}
=
\mathrm{BCE}(y_t,\hat y_t),
\qquad
y_t
=
\mathbb{I}\!\left[\mathcal{L}_{\mathrm{BC}}^{\mathrm{on}} < \mathcal{L}_{\mathrm{BC}}^{\mathrm{off}}\right].
\]
At inference time, we binarize \(\hat y_t\) to obtain \(\hat\alpha_t \in \{0,1\}\), which directly enables or disables the cross-attention message. We apply the same priors and regularizers from Sec.~\ref{Subsec:cooperation_learning} to \(\hat y_t\) as a soft relaxation, thereby shaping the resulting discrete gate.
This tokenized formulation simplifies gate prediction and empirically improved  stability in preliminary experiments. Implementation and training details are provided in appendix.


\section{Experiments}~\label{Sec: Experiments}
In this section, we evaluate our model’s ability to (i) freely reuse learned single-arm skills, (ii) capture the inter-arm coordination required by truly cooperative behaviors, and (iii) identify when cooperation is necessary for a given state. Our experiments are designed to answer the following questions:
\begin{enumerate}
    \item \label{exp:qn1} Can {SkillVLA} flexibly recompose learned skills and generalize to unseen combinations of single-arm skills?
    \item \label{exp:qn2} Is {SkillVLA}'s communication mechanism sufficient to support tight inter-arm coordination and reproduce cooperative dual-arm skills?
    \item \label{exp:qn3} Can {SkillVLA} handle multi-stage, long-horizon tasks by estimating cooperation level and switching between operating modes?
    \item \label{exp:qn4} In continual learning, does {SkillVLA} better leverage previously learned skills to acquire related new skills compared to existing methods?
\end{enumerate}

\mypara{Baselines and evaluation metrics.}
We compare {SkillVLA} against \(\pi_{0.5}\)~\cite{pi05}, a representative state-of-the-art VLA baseline. For Q\ref{exp:qn1}, we additionally include \(\pi_{0}\)-FAST~\cite{piFAST} as a representative autoregressive VLA, which represents the other major paradigm beyond diffusion-based VLAs and tests whether skill entanglement persists across architectures.
In the Q\ref{exp:qn1} setting, these general-purpose VLAs achieve near-zero success due to severe skill entanglement. To enable more informative comparisons, we further include TwinVLA~\cite{TwinVLA} as a factorized baseline that composes two single-arm VLAs by sharing action experts while allowing \emph{uncontrolled} cross-arm attention.
For most tasks (except the long-horizon task), we report \emph{success rate} as the primary metric. For long-horizon tasks which comprise multiple stages, we additionally report a \emph{progress score} and the \emph{average completion time} to quantify execution quality and efficiency. Additional experimental details are provided in appendix.

\subsection{Single-Arm Skill Reuse}

\mypara{Task Setup.} Here, we evaluate whether a bimanual VLA can \emph{reuse} learned single-arm skills under \emph{unseen} left--right combinations.
As shown in Fig.~\ref{Fig:exp_real_recomp}, our training data include three skills per arm, collected from single-arm executions while the other arm remains stationary.
At test time, we ask the policy to execute a left-arm skill and a right-arm skill \emph{simultaneously} in combinations that never appear in the training set.
This setting directly probes skill reuse since the policy should invoke the correct per-arm behaviors and maintain stable concurrent execution, rather than reverting to familiar demonstration-time patterns.

\mypara{Performance Results.} As reported in Tab.~\ref{Tab: exp_results_recompo} and Tab.~\ref{Tab: exp_results_skills_learned}, all methods perform comparably on the \emph{seen} single-arm skills, but their behaviors \textit{diverge sharply} when asked to \emph{zero-shot} recompose unseen left-right combinations.

In particular, \(\pi_{0.5}\) and \(\pi_0\)-FAST achieve a \(0\%\) success rate.
Qualitatively, these models frequently revert to the training-time regime in which one arm acts while the other remains idle, indicating that they primarily reproduce skill pairings present in the dataset rather than freely recombining per-arm behaviors. 
Moreover, when instructed to perform a two-arm task, \(\pi_{0.5}\) often exhibits instability in that the nominally ``idle'' arm jitters or drifts. These results are not entirely surprising since the test settings are essentially out-of-distribution (OOD) for standard VLAs and this failure mode can be attributed to \emph{entangled} bimanual control.
Because \(\pi_{0.5}\) uses a single shared representation to control both arms, each left--right pairing effectively corresponds to a distinct latent mode specifying a coupled bimanual action.
Unseen pairings therefore require a latent that is not induced by any training demonstration, producing an OOD input to the downstream action expert. 

TwinVLA attains non-zero success by occasionally producing the intended composition. However, although it uses separate VLM streams for the two arms, it relies on \emph{uncontrolled} cross-attention between streams, which can introduce implicit coupling.
This yields unpredictable interference under unseen pairings and keeps overall success low.

In contrast, {SkillVLA} achieves a substantially higher success rate of \(51\%\) on unseen combinations.
By explicitly disentangling per-arm skill invocation, performing skill-adaptive action generation, and enabling \emph{controlled} inter-arm communication only when needed, {SkillVLA} more reliably reuses the same single-arm behaviors under novel pairings while reducing spurious cross-arm interference.

\begin{figure}[t]
    \centering
    \includegraphics[width=0.90\linewidth]{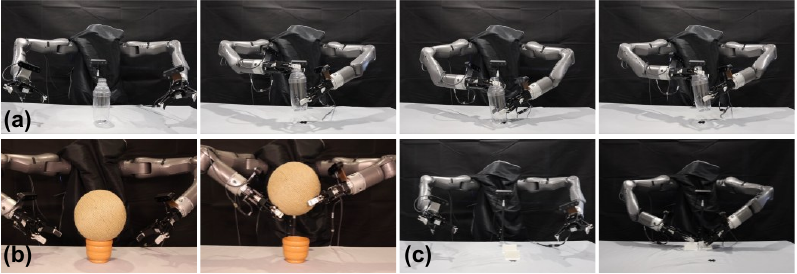}
    \caption{\small \textbf{Cooperative tasks.} \textbf{(a)} \textsc{Shake}: Shake the cup with a cap without making them fall apart. \textbf{(b)} \textsc{Ball}: Lift the ball steadily. \textbf{(c)} \textsc{Align}: Align the blocks on the table.}
    \label{Fig:exp_cooperative_real}
\end{figure}

\begin{figure*}[t]
    \centering
    \includegraphics[width=0.90\linewidth]{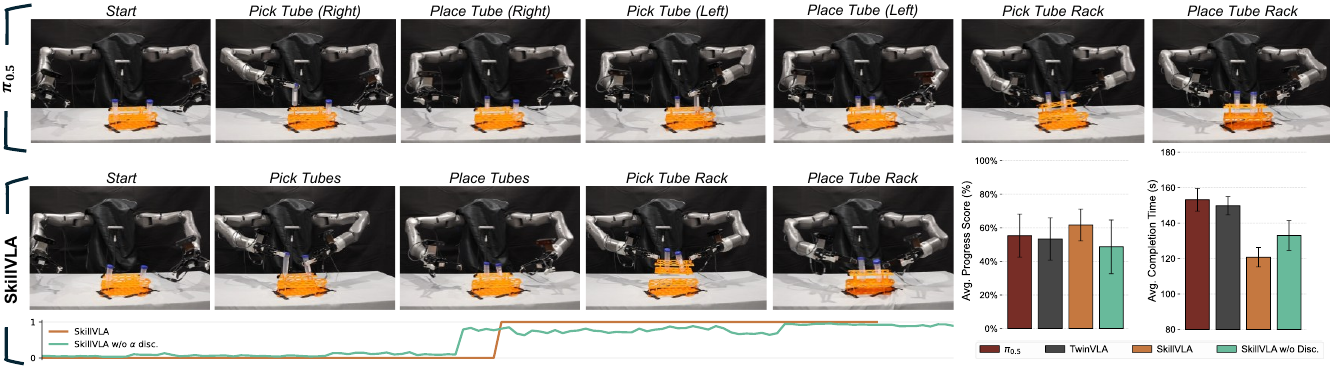}
    \caption{\small\textbf{Long-horizon tasks behaviors and results.} \textbf{Top:} Behavior of $\pi_{0.5}$ in \textsc{Tubes}. \textbf{Middle left:} Behavior of SkillVLA in \textsc{Tubes}. \textbf{Bottom left:} Changes of $\alpha$ values throughout the completion, respectively from SkillVLA and an ablated version without discretization of $\alpha$.
    \textbf{Bottom right:} Averaged progress score and completion time of methods on the long-horizon tasks.}
    \label{fig:exp_long_hor}
    \vspace{-2mm}
\end{figure*}
\begin{figure*}[t]
    \centering
    \includegraphics[width=0.90\linewidth]{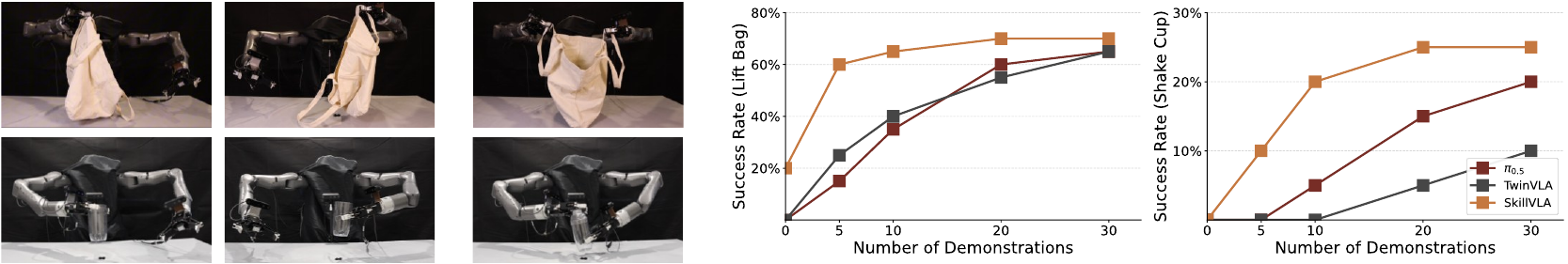}
    \caption{\small\textbf{Continual learning tasks and results.} \textbf{Left:} SkillVLA in continual learning tasks, \textsc{Lift bag} and \textsc{Shake cup}. \textbf{Right:} Success rates on cooperative tasks \textsc{Lift bag} and \textsc{Shake cup} w.r.t the number of continual learning demonstrations.}
    \label{fig:continual_learning}
    \vspace{-5mm}
\end{figure*}

\subsection{Dual-Arm (Cooperative) Skill Reproduction}

\mypara{Task Setup.} To evaluate whether {SkillVLA} can reproduce cooperative dual-arm skills that require close inter-arm coordination, we design three cooperative tasks (Fig.~\ref{Fig:exp_cooperative_real}). These tasks fall into two categories with qualitatively different coordination demands. The first category, \textsc{Shake} and \textsc{Ball}, requires \emph{tight low-level alignment} between the two arms. The second category, \textsc{Align}, requires both low-level alignment and \emph{high-level coordination}; two blocks are placed near the center of the table, and the two arms must simultaneously pick the objects and align them. Importantly, the arm-object assignment is randomized across trials, requiring the policy to resolve task allocation on the fly.

\mypara{Performance Results.} As reported in Tab.~\ref{tab:exp_cooperative}, {SkillVLA} achieves performance comparable to the strong baseline \(\pi_{0.5}\) across all cooperative tasks. This suggests that {SkillVLA} can identify when cooperation is required and that its explicit communication mechanism is expressive enough to support tight inter-arm coordination when needed.

\mypara{Ablation: Cross-Attention.} We further isolate the role of action-level communication by evaluating an ablation that disables cross-attention between the two action experts. This variant performs particularly poorly on \textsc{Shake} and \textsc{Ball}, confirming that cross-attention-based communication is important for maintaining tight low-level alignment. On \textsc{Align}, which places greater emphasis on high-level task assignment and is comparatively less sensitive to continuous low-level coupling, disabling communication results in a smaller performance drop.

\subsection{Applications in Long-Horizon Settings}
We next evaluate whether {SkillVLA} can (i) infer the required cooperation level and (ii) switch between collaborative and non-collaborative modes \emph{within a single episode}. 
\begin{table}
  \centering
  \caption{\small Success Rate in Cooperative Tasks $\uparrow$}
  \label{tab:exp_cooperative}
  \resizebox{0.98\linewidth}{!}{%
  \begin{threeparttable}
    \begin{tabular}{lcccc}
      \toprule
      \textbf{Methods} & \textbf{Shake} & \textbf{Ball} &
      \textbf{Align} & \textbf{Avg.} \\
      \midrule
      $\pi_{0.5}$  & \textbf{0.30} & 0.45 & 0.65 & 0.47 \\
      TwinVLA      & 0.15 & \textbf{0.55} & 0.55 & 0.42 \\
      SkillVLA   & 0.25 & 0.50 & \textbf{0.70} & \textbf{0.48} \\
      SkillVLA w/o Attn.  & 0.00 & 0.10 & 0.40 & 0.17 \\
      \bottomrule
    \end{tabular}
    \vspace{-3mm}
  \end{threeparttable}
  }
\end{table}
\mypara{Task setup.} To this end, we design two long-horizon real-world tasks, \textsc{Tubes} and \textsc{Collect Items} (Fig.~\ref{fig:exp_long_hor}). Each task intentionally interleaves phases that admit independent per-arm execution with phases that require explicit bimanual cooperation. As a representative example, \textsc{Tubes} consists of two stages with distinct coordination demands:
(1) transferring tubes from one rack to another (independent per-arm collection), and
(2) lifting and placing the loaded rack onto another rack (inherently cooperative bimanual manipulation).
We train the model using demonstrations of three skills, i.e., transferring a tube with the left arm, transferring a tube with the right arm, and moving the rack with both arms. We then evaluate the policy \emph{zero-shot} on the full long-horizon task that composes these stages. 

\mypara{Performance and Efficiency Results.}
As shown in Fig.~\ref{fig:exp_long_hor} (bottom right), {SkillVLA} achieves a progress score comparable to strong baselines, indicating that it can select an appropriate operating mode and complete the task reliably. More importantly, {SkillVLA} \textit{reduces average completion time} by \(\approx 21\%\) relative to the baselines. Qualitatively, monolithic baselines tend to follow demonstration-like sequential patterns (one arm acts while the other waits), whereas \textbf{SkillVLA} leverages \emph{skill reuse} to exploit parallelism when available. In \textsc{Tubes}, during the tube-collection stage, it concurrently invokes the single-arm tube-transfer skills with both arms (i.e., \(\alpha=0\)), enabling parallel left--right execution and substantially accelerating completion. This ability to opportunistically parallelize subtasks better utilizes the robot’s hardware and improves practical throughput.

\mypara{Mode switching via the cooperation gate.}
Fig.~\ref{fig:exp_long_hor} (bottom left) visualizes \textbf{SkillVLA}'s discrete gate output \(\alpha \in \{0,1\}\), where \(\alpha=0\) corresponds to independent per-arm execution and \(\alpha=1\) activates inter-arm cooperation. We observe that \(\alpha\) evolves coherently with task progress; it switches to \(\alpha=1\) during inherently cooperative phases such as moving the loaded rack with both arms, while remaining at \(\alpha=0\) during collection phases where the arms can operate independently. This stage-consistent switching suggests that the gate can infer the required cooperation level from scene and instruction context, allowing {SkillVLA} to reuse both single-arm and dual-arm behaviors within a single rollout.

\mypara{Ablation: Continuous vs.\ Discretized Gating.}
We also evaluate an ablation that uses a continuous (non-discretized) gate. This variant is less stable over long horizons. As shown in Fig.~\ref{fig:exp_long_hor} (bottom left), continuous \(\alpha\) exhibits noticeably higher fluctuation despite following a similar overall trend. This instability propagates to action generation, leading to lower success and higher variance in performance (Fig.~\ref{fig:exp_long_hor}, bottom right). These results suggest that discretizing \(\alpha\) improves robustness by enforcing simpler, clearer mode switching.

\subsection{Reuse of Learned Skills in Continual Learning}
\mypara{Task setup.} 
To answer Q\ref{exp:qn4}, we study whether SkillVLA can reuse learned single-arm skills to accelerate the acquisition of a new cooperative dual-arm skill via fine-tuning. As illustrated in Fig.~\ref{fig:continual_learning} (left), we consider two task groups, \textsc{Lift bag} and \textsc{Shake cup}. In each group, we first pretrain the model on the corresponding single-arm dataset (e.g., lifting a bag with one arm), and then fine-tune it with a small number of demonstrations to obtain the target dual-arm skill (e.g., lifting with both arms).
For each task group, we construct a two-stage continual-learning protocol: (i) pretraining on single-arm executions to learn reusable per-arm behaviors, followed by (ii) fine-tuning on a limited set of dual-arm demonstrations to learn the cooperative variant. 

\mypara{Learning Curves and Data Efficiency.}
Fig.~\ref{fig:continual_learning} (right) reports success rates as a function of the number of dual-arm demonstrations used for fine-tuning. In \textsc{Lift bag}, the target dual-arm behavior is close to a composition of the pretrained single-arm skills, and SkillVLA achieves \(20\%\) success \emph{zero-shot}. Moreover, with only 5 fine-tuning demonstrations, it reaches near-peak performance. In contrast, \(\pi_{0.5}\) and TwinVLA remain far from convergence even after training on 30 demonstrations. These results indicate that SkillVLA can effectively leverage previously learned single-arm behaviors for rapid adaptation.

\mypara{Adapting Skills to New Coordination.}
\textsc{Shake cup} is challenging because the required coordination deviates from the original single-arm behavior; one hand must stabilize the cap rather than simply shaking the cup body. Despite this mismatch, SkillVLA exhibits the same data-efficient trend. This suggests that reuse in continual learning is not limited to replaying identical action patterns. Instead, the model can transfer related single-arm behaviors as a strong initialization and quickly adapt them into a new cooperative policy. Overall, these results support skill reuse as a practical mechanism for continual learning and for scaling generalist bimanual policies with reduced demonstration burden.

\section{Conclusion, Limitations, and Future Work}~\label{Sec:limitation}
In this work, we study the \emph{combinatorial diversity} challenge in bimanual manipulation through the lens of \emph{skill reuse}. We identify \emph{skill entanglement} in existing VLA designs as a central obstacle to reuse---particularly for composing diverse left--right skill pairings and generalizing to unseen task combinations. To address this, we propose {SkillVLA}, which mitigates entanglement by explicitly separating per-arm skill invocation and enabling skill-adaptive action generation, while still leveraging the generalization of pretrained vision--language models. Across real-world evaluations, SkillVLA improves combinatorial generalization and data efficiency in continual skill acquisition, demonstrating more reliable reuse of both single-arm and cooperative dual-arm behaviors.

\mypara{Limitations.} SkillVLA depends on pretrained VLMs for high-level reasoning and for providing priors on cooperation, which ties performance to the capability, latency, and cost profile of the underlying foundation model. Future work will explore more efficient ways to distill or compress this reasoning into lightweight policy components, as well as richer skill representations beyond natural language. More broadly, we view disentangled skill learning/selection and controlled cross-arm coupling as key ingredients for scaling toward truly generalist bimanual agents that can recombine skills reliably on demand.





\balance
\bibliographystyle{plainnat}
\bibliography{references}

\clearpage 
\nobalance 

\clearpage
\appendices
\section{Method Analysis} 
In this appendix, we provide a comparative analysis to justify the architectural design of SkillVLA. We identify \textit{skill entanglement} in existing monolithic VLAs as a fundamental obstacle to combinatorial generalization. Building on these structural findings, we demonstrate the necessity of the fully disentangled approach adopted by SkillVLA and analyze its effectiveness in resolving entanglement. We begin by establishing the geometric definitions that underpin our discussion.

\mypara{Preliminary: Support.}
To facilitate our geometric analysis, we first establish the concept of \textit{effective support} to delineate the valid behavior of a learned policy. For a continuous policy distribution $\pi(\cdot|x)$, we define the effective support as the set of actions where the probability density is non-negligible:
$$
\text{supp}(\pi(\cdot|x)) = \{ a \in \mathcal{A} \mid \pi(a|x) > \epsilon \}
$$
In this analysis, we use the support to distinguish between the \textit{training manifold} (the region of the action space covered by expert demonstrations) and the \textit{target support} (the region of the action space required to solve a novel recomposition task). A model's ability to generalize to unseen tasks depends on whether its predicted support can cover the target support of the new context.

\subsection{Action Entanglement in Monolithic VLAs}

In this section, we provide a geometric analysis of the limitations inherent to monolithic bimanual VLAs. We demonstrate that modeling dual-arm control via a single joint distribution $\pi_{\theta}(a_L, a_R | x)$ creates a structural disconnect between the training manifold and the target space required for skill recomposition. We term this limitation \textit{Action Entanglement}.

\mypara{Formulation of the Training Manifold. }Consider a monolithic VLA policy trained on a dataset of expert demonstrations $\mathcal{D}$. We define the \textit{training manifold} $\mathcal{S}_{train}$ as the subset of the joint action space supported by these demonstrations:
$$
\mathcal{S}_{train} = \{ (a_L, a_R) \mid \exists x, (x, a_L, a_R) \in \mathcal{D} \} \subset \mathcal{A}_L \times \mathcal{A}_R
$$
In standard bimanual datasets, this manifold is typically sparse and exhibits high inter-arm correlation. For instance, if a specific ``lifting'' skill for the left arm is only ever demonstrated while the right arm is ``idling'', the manifold $\mathcal{S}_{train}$ captures this specific coupling. Consequently, the learned policy $\pi_{\theta}$ approximates the empirical joint density defined strictly over this support, internalizing the conditional dependence between the arms.

\mypara{The Geometry of Unseen Recomposition.} Consider a recomposition task defined by a context $x^*$ that requires the concurrent execution of a known left-arm skill $s_L$ and a known right-arm skill $s_R$. The set of valid solutions, or the \textit{target support}, is the Cartesian product of the valid marginal actions for each skill:

$$
\mathcal{S}_{target}(x^*) = \text{supp}(\pi_{s_L}(\cdot|x)) \times \text{supp}(\pi_{s_R}(\cdot|x))
$$

We define a skill combination as ``unseen'' when this specific pairing of behaviors has not been observed in the dataset. Geometrically, this implies that the target solution lies outside the training manifold:
$$
\mathcal{S}_{target}(x^*) \cap \mathcal{S}_{train} = \emptyset
$$

\mypara{Entanglement Mechanism.} The core challenge of action entanglement arises from the discrepancy between the training objective and the inference requirement under this disjoint geometry.

\textit{Training Dynamics:} The objective of imitation learning is to maximize the likelihood of actions within $\mathcal{S}_{train}$. The policy $\pi_{\theta}$ inherently learns the correlations present in the data (e.g., $a_L \in \text{Lift} \implies a_R \in \text{Idle}$).

\textit{Inference Failure:} When presented with $x^*$, the model is required to generate an action $a^* \in \mathcal{S}_{target}$. However, because $a^*$ lies in a region of the joint space assigned negligible probability density by the learned distribution (since $a^* \notin \mathcal{S}_{train}$), the monolithic model struggles to factorize the arms. Instead, the policy tends to regress to the high-density modes of the training distribution. This results in spurious crosstalk, where the execution of one arm creates a ``pull'' that forces the other arm to default to its training-time partner behavior, preventing the successful composition of the two skills.

\subsection{Latent Entanglement in Action-Expert VLAs}

We consider the broad class of action-expert VLAs, where a pretrained VLM backbone encodes the multimodal context $x$ into a shared latent representation $z$, and a downstream action module predicts actions conditioned on this $z$. While standard implementations typically predict the joint action $(a_L, a_R)$ monolithically (thus suffering from action entanglement), one might propose a straightforward solution: factorizing the output module to use disjoint action experts $\pi_L(a_L|z)$ and $\pi_R(a_R|z)$.

In this section, we demonstrate that this decoupled architecture remains insufficient for reliable recomposition due to a secondary failure mode at the vision-language interface, which we term \textit{Latent Entanglement}.

\mypara{Geometry of the Latent Interface. }During training, the VLM backbone projects the demonstration inputs $\mathcal{D}$ onto a specific manifold within the latent space. We denote the effective support of these training latents as:
$$
\mathcal{Z}_{train} = \{ z = f_{\text{VLM}}(x) \mid x \in \mathcal{D} \}
$$
The downstream action experts (AEs), $\pi_L$ and $\pi_R$, are optimized to approximate the conditional action density $p(a|z)$ specifically for inputs $z \in \mathcal{Z}_{train}$. Consequently, the reliability of these experts is strictly bounded by this learned latent support.

\mypara{Generalization Failure via OOD Latents. }Consider an unseen recomposition task $x^*$ requiring a novel combination of skills. While the high-level VLM may generally understand this scene, it projects the global context into a single embedding $z^* = f_{\text{VLM}}(x^*)$.

Because $x^*$ represents a semantic combination distinct from any training example, the resulting embedding $z^*$ may deviate from the cluster of latents seen during training (i.e., $z^* \notin \mathcal{Z}_{train}$). This presents the downstream experts with an Out-of-Distribution (OOD) conditioning variable. Unlike the VLM backbone, these specialized experts lack web-scale pretraining and are prone to unpredictable behavior when queried with latents outside their training manifold.

\mypara{Conclusion: The Entanglement of Reasoning.} This analysis reveals that the reliance on a single shared latent creates a bottleneck. The latent $z$ must implicitly encode the cross-arm dependencies observed in the data to satisfy the training objective. Consequently, even if the action heads are structurally disentangled, the shared reasoning pathway entangles the control signals before they reach the experts.

This highlights that true combinatorial generalization requires more than just output factorization; it requires the \textit{disentanglement of reasoning}. The VLM must be capable of explicitly decomposing the global context into separate skill descriptors for each arm, rather than summarizing the joint state into a single entangled embedding.

\subsection{Analysis of SkillVLA}

Our theoretical analysis has identified action and latent entanglement as critical barriers to skill reuse, arising from the mismatch between the training manifold and the target space of unseen tasks. SkillVLA is explicitly designed to resolve this mismatch. In this section, we analyze the generative process of our method, demonstrating how it structurally supports single-arm skill recomposition where monolithic baselines fail.

When the collaboration gate is closed ($\alpha=0$), SkillVLA models the joint action distribution as a conditional product of independent streams:
\begin{align*}
\pi_\psi(a_L, a_R &\mid x) = \int_{u, z} \underbrace{p(a_L \mid z_L) p(a_R \mid z_R)}_{\text{Independent AEs}} \\ &\cdot\underbrace{p(z_L \mid x, u_L) p(z_R \mid x, u_R)}_{\text{Per-Arm VLMs}} \cdot\underbrace{p(u_L, u_R \mid x)}_{\text{High-level VLM}} du dz.
\end{align*}
This factorization enables the model to effectively cover the target support $\mathcal{S}_{target}$ for unseen skill pairs. We analyze this process in three stages:

\mypara{Semantic Decomposition to Known Primitives}
First, leveraging its web-scale generalization capability, the high-level planner decomposes the novel context $x^*$ into explicit skill descriptors $u_L$ and $u_R$. Crucially, while the \textit{combination} $(u_L, u_R)$ is novel to the system, the individual descriptors correspond to single-arm skills $s_L$ and $s_R$ that exist within the training repertoire. This step effectively breaks the OOD joint query into recognizable semantic units.

\mypara{In-Distribution Latent Generation}
Unlike shared-latent models that must encode the novel combination into a single vector, SkillVLA processes the descriptors independently: $z_L \sim p(z_L | x, u_L)$. Since $u_L$ specifies a known skill $s_L$ present in the training demonstrations $\mathcal{D}$, the mapping from $(x, u_L)$ to $z_L$ constitutes an in-distribution inference task for the low-level VLM. Consequently, the generated latent $z_L$ naturally aligns with the manifold of latents seen during training ($\mathcal{Z}_{train}$), preventing the drift into undefined regions that characterizes latent entanglement.

\mypara{Construction of the Product Support}
With valid, in-distribution latents $z_L$ and $z_R$, the independent Action Experts $p(a_L | z_L)$ and $p(a_R | z_R)$ operate within their valid domains. The resulting joint policy support is the Cartesian product of these valid marginals, 
$$
\text{supp}(\pi_{\psi}) = \text{supp}(p(a_L | z_L)) \times \text{supp}(p(a_R | z_R))
$$
Since the experts correctly recover the behaviors for the known skills $s_L$ and $s_R$, this product space effectively covers the target support $\mathcal{S}_{target}(x^*)$. Thus, by structurally reducing an OOD joint task into ID sub-tasks, SkillVLA provides a robust mechanism for zero-shot skill recomposition.
\section{Implementation Details}
\label{Appendix: imple}
\mypara{Methods Implementation.} Our goal is to isolate the structural limitations of existing VLA systems and evaluate our proposed remedy under controlled settings. To avoid confounding factors, we do not introduce new backbone architectures or extensive pretraining recipes. Instead, we adopt minimal structural modifications to the primary baseline $\pi_{0.5}$~\cite{pi05}, ensuring the pipeline aligns with our disentangled design while maintaining component comparability.

We initialize the components using the pretrained PaliGemma checkpoints released with $\pi_{0.5}$~\cite{pi05}, duplicating the low-level VLM and action experts to support disentangled execution. While directly using an off-the-shelf VLM is possible, prior to policy learning, we perform a lightweight vision-language fine-tuning of the high-level VLM on subtask generation data to ensure prompting stability; this module then remains frozen throughout the main training phase, where only the low-level VLMs are fine-tuned via independent LoRA adapters~\cite{lora}.

The cross-attention module, which enables controlled interaction between the two action experts, employs independent query/key/value (QKV) projections trained from scratch. For action sampling, we maintain the flow-matching time schedule identical to the original $\pi_{0.5}$ implementation. Finally, the cooperation-level estimator is implemented as a Transformer decoder that attends to the high-level VLM's KV cache via cross-attention.

\mypara{Cooperation Prior.} To obtain the cooperation level labels used as learning priors, we preprocess the training dataset using a frozen Qwen3-VL-32B model~\cite{qwen3vl}. Specifically, we prompt the model annotate the degree of required inter-arm coordination, augmenting the dataset with these values as part of ground-truth supervision for the cooperation estimator.
\begin{table*}[t]
\centering
\caption{Cooperative Task Details}
\label{appendix_tab:coop_task_list}
\setlength{\tabcolsep}{4pt}
\renewcommand{\arraystretch}{1.1}
\begin{tabularx}{\textwidth}{l l X}
\toprule
\textbf{Abbr.} & \textbf{Task} & \textbf{Description}\\
\midrule
Shake & \textsc{Shake Cup} & A cup with a loose cap is placed on the table. The robot must shake the cup bimanually, with one arm securing the cap to prevent separation. This requires the arms to move synchronously to maintain the assembly's integrity. \\
Lift & \textsc{Lift Ball} & A ball rests on a stand. The robot must lift the ball using both arms. The upward motion must be synchronized in velocity to maintain equilibrium and prevent the ball from rotating or falling. \\
Align & \textsc{Align Blocks} & Two blocks are placed on the table. The robot must grasp one block with each arm and place them onto the table simultaneously such that they align collinearly. \\
\bottomrule
\end{tabularx}
\end{table*}
\begin{figure*}[t]
    \centering
    \includegraphics[width=0.98\linewidth]{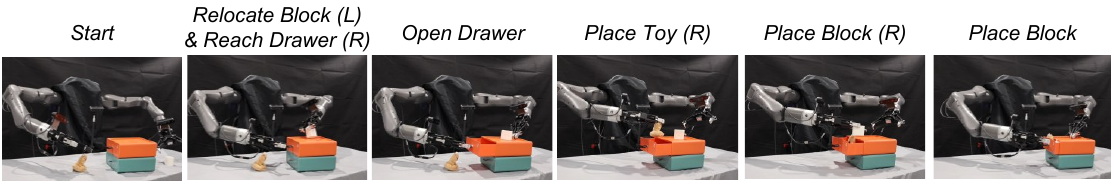}
    \caption{Rollout of SkillVLA on \textsc{Collect Items}.}
    \label{appFig: exp_collect_item}
\end{figure*}

\section{Experimental Details}
\label{app: exps}
\subsection{Baselines}
\mypara{$\boldsymbol{\pi_0}$\textbf{-FAST and }$\boldsymbol{\pi_{0.5}}$\textbf{. }} Contemporary VLA architectures predominantly fall into two paradigms: autoregressive generation and diffusion-based modeling. We select $\pi_0$-FAST~\cite{piFAST} and $\pi_{0.5}$~\cite{pi05} as representative state-of-the-art baselines for these respective categories, allowing us to evaluate performance across both dominant design spaces.

\mypara{TwinVLA.} To extend our evaluation to architectures beyond mainstream monolithic models, we additionally include TwinVLA~\cite{TwinVLA} as a baseline. This method is constructed by duplicating a single-arm VLA to improve training efficiency, incorporating cross-attention between the parallel VLM streams to enable inter-arm communication. We reproduce this architecture to evaluate the performance of current factorized designs.
\subsection{Experimental Details}
\label{app: exp_details}
\subsubsection{Skill Recomposition Tasks}
We conduct skill recomposition experiments to evaluate the capability for single-arm skill reuse. In our protocol, models are trained on a set of independent single-arm skills, using 50 demonstrations per skill. We first validate that the models have mastered these source behaviors by conducting 10 evaluation rollouts for each original skill. Subsequently, we assess zero-shot generalization by testing the models on unseen skill combinations, performing 10 evaluation episodes for each pairing to establish the comparative trends in skill reuse. 
\subsubsection{Cooperative Tasks}
Our experiments include three cooperative tasks—\textsc{Shake Cup}, \textsc{Lift Ball}, and \textsc{Align Blocks}—designed to probe the efficacy of inter-arm communication (see Table~\ref{appendix_tab:coop_task_list} for details). Each task necessitates strict bimanual coupling, ranging from synchronized low-level actuation to high-level planning. For each distinct task, we collect a dataset of 50 demonstrations and train task-specific models. To ensure statistical reliability for evaluating performances, we evaluate each model on each task over 20 rollouts.
\begin{figure*}[t]
    \centering
    \includegraphics[width=0.98\linewidth]{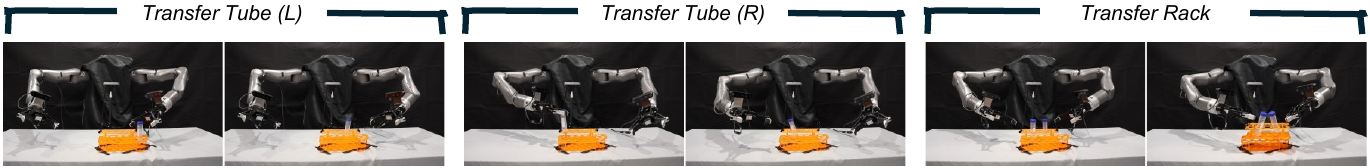}
    \caption{Skills Learned for \textsc{Tubes}.}
    \label{appFig: exp_tubes_skills}
\end{figure*}
\begin{figure*}[t]
    \centering
    \includegraphics[width=0.98\linewidth]{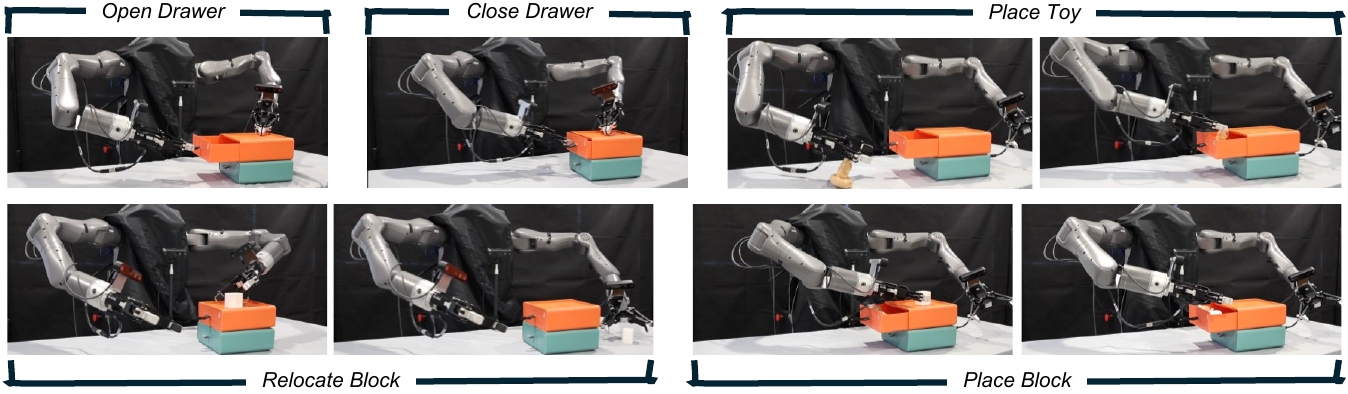}
    \caption{Skills Learned for \textsc{Collect Items}.}
    \label{appFig: exp_collect_item_skills}
\end{figure*}
\subsubsection{Long-Horizon Tasks}
\label{app: exp_details_long_horizon}
Our experiments include two long-horizon tasks: \textsc{Tubes} and \textsc{Collect Items}. We train task-specific policies separately using demonstrations of the skills required for each task (Fig.~\ref{appFig: exp_tubes_skills} and \ref{appFig: exp_collect_item_skills}), collecting 50 demonstrations per skill. For evaluation, we use the same set of items in the same environment, while randomizing the initial object placements across trials to ensure a fair comparison. We run 10 evaluation episodes per task, resulting in 20 total evaluations for each method.

For evaluation, we report the \emph{average progress rate} to reflect overall task performance. The progress rate of each episode is defined as the achieved progress score divided by the maximum score of the corresponding task. Details for each task are as follows:
\begin{itemize}
    \item \textsc{Tubes:} The task starts with two tube racks placed next to each other on the table: one rack is empty and the other holds two tubes. The robot must (i) transfer both tubes to the empty rack, and then (ii) lift the rack containing the tubes using both arms and place it onto the other rack.
    \begin{itemize}
        \item[+1] For each tube successfully transferred. $\times2$
        \item[+1] For successfully placing the rack.
    \end{itemize}
    \textit{Maximum score: 3 points.}

    \item \textsc{Collect Items:} The task starts with two items and a closed drawer on the table. The robot must (i) open the drawer using both arms, (ii) load both items into the drawer, and then (iii) close the drawer using both arms. Notably, as shown in Fig.~\ref{appFig: exp_collect_item}, one item is initially located behind the drawer and cannot be placed into it directly. The robot must first use one arm to relocate this item to a position above the drawer reachable to the other arm, and then use the other arm to place it into the drawer.
    \begin{itemize}
        \item[+1] For opening the drawer.
        \item[+1] For relocating the behind-drawer item to above the drawer.
        \item[+1] For each item placed into the drawer. $\times2$
        \item[+1] For closing the drawer.
    \end{itemize}
    \textit{Maximum score: 5 points.}
\end{itemize}

Additionally, we report completion time to reflect behavioral efficiency. Since fully successful episodes are sparse, we compute a \emph{progress-normalized completion time} and average it only over episodes with non-zero progress. Concretely,
\[
t_{\text{norm}}=\frac{1}{|\mathcal{I}|}\sum_{i\in\mathcal{I}} \frac{t_i}{s_i},
\]
where $t_i$ is the execution time of the $i$-th evaluation episode, $s_i\in(0,1]$ is its progress rate, and $|\mathcal{I}|$ denotes the number of episodes that achieve non-zero progress.
\subsubsection{Continual-Learning Tasks}
Our evaluation includes two groups of continual learning tasks. In this protocol, we first pretrain the models on fundamental single-arm skills using 50 demonstrations per skill. Subsequently, we fine-tune the models on dual-arm tasks using varying numbers of limited demonstrations, as stated in the main text. To ensure statistical reliability, we evaluate the models over 20 rollouts after each fine-tuning stage.

\end{document}